# *AMOSL:* ADAPTIVE MODALITY-WISE STRUCTURE LEARNING IN MULTI-VIEW GRAPH NEURAL NETWORKS FOR ENHANCED UNIFIED REPRESENTATION

Peiyu Liang[1], Hongchang Gao, and Xubin He

Department of Computer and Information Sciences, Temple University, Philadelphia, PA, USA, 19122

## ABSTRACT

*While Multi-view Graph Neural Networks (MVGNNs) excel at leveraging diverse modalities for learning object representation, existing methods assume identical local topology structures across modalities that overlook real-world discrepancies. This leads MVGNNs struggles in modality fusion and representations denoising. To address these issues, we propose adaptive modality-wise structure learning (AMoSL). AMoSL captures node correspondences between modalities via optimal transport, and jointly learning with graph embedding. To enable efficient end-to-end training, we employ an efficient solution for the resulting complex bilevel optimization problem. Furthermore, AMoSL adapts to downstream tasks through unsupervised learning on inter-modality distances. The effectiveness of AMoSL is demonstrated by its ability to train more accurate graph classifiers on six benchmark datasets.*

## 1. INTRODUCTION

A large amount of data networks exhibits a unique structure known as graph-structured data. Irregular, non-Euclidean data characterize this data type and are frequently found in areas such as recommendation systems [1], social media networks [2], knowledge graphs [3], and molecular structures [4]. The analysis of graph-structured data has garnered substantial attention for its inherent inductive and transductive properties, which enable relational reasoning among entities (nodes) and their connections (edges).

Graph Neural Networks (GNNs), through a series of advancements [5][6][7][8][9][10][11][12][13], have shown promising improvement in studying graph-structured data to conduct various downstream graph- related tasks. These tasks include node classification, link prediction, and graph classification. GNNs inherit key operations from Convolutional Neural Networks (CNNs), including convolution and pooling, while leveraging layer-by-layer graph signal processing [14] and message passing [15] to capture inherent structures within the graphs effectively.

Drawing from the capabilities of GNNs, Multi-view Graph Neural Networks (MVGNNs) are designed to learn multiple input modalities, each providing a unique perspective to understand the same underlying object. These networks make use of parallel and independent GNNs to capture modality-specific representations, which are subsequently fused into a unified representation. This framework positions MVGNNs for potential superiority over single-view GNNs and is supported by practical applications across diverse domains. Examples include the identification of molecular networks characterized by diverse underlying molecular structures [16], prediction of brain diseases using various neuroimaging data sources [17], analysis of global poverty patterns across different countries [18], and analysis of user activities on social networks from their both online and offline interactions [19].

MVGNNs, while effective, face several noteworthy challenges: 1) Labeling variability: Modalities could fall apart for being labeled to different classes in a given feature space, due to parallel learning with unshared GNN parameters. Fusing these representations from disparate classes can significantly increase the potential for noise in the resulting unified representation; 2) Node correspondence ambiguity: Different modalities often exhibit distinct graph topologies, making element-wise fusion challenging due to unclear node correspondences across modalities. As an example, modality-wise pooling, a commonly used fusion technique, involves selecting specific merits from modality-specific representations on an element-wise basis. However, without establishing clear node correspondences before fusion, information underutilization and noise in the unified representation can occur; 3) Inter-modality correlation neglect and knowledge underutilization: Modalities should inherently exhibit some level of correlation since they all pertain to a single object. However, neglecting to capture the inter-modality correlation can impede the utilization of the complementary knowledge carried by diverse modalities.

Optimal transport distance, alternatively known as Wasserstein distance [20] in Mathematics or Earth Mover's Distance [21] in Computer Science, is a linear programming (LP) problem. Its objective is to minimize transportation costs between two probability distributions while identifying transport correspondences and quantifying the transport flow between these correspondences. This concept extends to various domains, including graph learning, where it applies to probability distributions such as graph filters [22] or graph signals [23]. Recently, extensive research efforts have delved into the graph problem through the perspective of optimal transport distance. These studies span various domains, including graph matching [22][24][25][26] and node classification [27][23]. Indeed, these studies have predominantly contributed to the advancements in graph-based machine learning, often centered around off-the-shelf machine learning models or single-layer neural networks. However, incorporating the optimal transport distance into deep neural networks still needs to be investigated in problem formulation and method optimization.

In this paper, we introduce an optimal transport-based adaptive modality-wise structural learning (AMoSL) in MVGNNs to address challenges specific to MVGNNs:

1. To mitigate the issue of modalities potentially falling into different class spaces due to parallel learning with unshared GNN parameters, we propose an unsupervised learning method that focuses on the distances between modalities.

2. To address the ambiguity in node correspondence and improve the fusion process while reducing noise in the unified representation, we explore node correspondence among modalities using an optimal transport metric.

3. To tackle the neglect of inter-modality correlations and knowledge underutilization, we adopt a dual approach involving node-level and graph-level considerations. At the node level, we see corresponding nodes as inter-modality correlations, intending to minimize the distance between these nodes to preserve shared knowledge. At the graph level, in addition to unsupervised minimization of inter-modality distances, we guide the model to maintain balance by adapting to the classifier's performance. This prevents modalities from diverging too far, neglecting shared knowledge, or converging too closely, underutilizing complementary knowledge.

Our method demonstrates its effectiveness by achieving improved graph classification results on six benchmark datasets. We also conduct experimental evaluations and ablation studies to further validate our approach.

## 2. RELATED WORK

**Multi-view Graph Neural Networks.** MVGNNs are built for modeling and relation reasoning among graphs belonging to the same object in different views. Inspired by a study in CNNs [28], which projected a 3D object onto 2D images to better recognize the object from different views, in MVGNNs, input graphs in different views are multi-structured or multi-related data. For example, [17][29][30] used different tractography methods in brain network studies, [18] used different data resources from different developing countries in a global poverty study, and [16] artificially created hybrid modalities to study data in the complex structure. To utilize and combine the learned knowledge from different modalities, [17] developed a voting strategy to balance the output from different graph signals, [30] used a Hadamard product layer to capture graphs similarity, [29] concatenated graph-level knowledge to preserve all knowledge from modalities, and [31] regularized the distance between two modalities for node-level classification. However, most existing MVGNNs neglect the inter-modality correlation.

**Optimal Transport.** In a study of content-based image retrieval [21], optimal transport (OT) proves its certain robustness in partial matching and flexibility in handling variable-length signals than in histogram matching methods. It is the first work for OT involved in the neural network, followed by further applications across various domains, particularly in documents and images. For example, in document classification [32], OT optimizes the transfer cost of word embedding from one document to another. In image classification, OT overcomes the issue of the limited amount of labeled training data to achieve few-shot learning [33]. Recently, OT has gained significant attention in node classification tasks related to graph alignment [22][27][23]. OT in this domain involves aligning corresponding nodes or topologies between graphs for unlabeled node prediction. Because of the optimization challenges, these studies, are either off-the-shelf machine learning models or single-layer neural networks. In this work, we overcome the optimization challenge of integrating OT into deep Graph Neural Networks and apply its properties to improve graph-level classification in MVGNNs.

**Parameterized Optimization Learning.** Parameterized optimization learnings typically apply to supervised machine learning and follow differentiation through argmin techniques [34][35][36][37][38]. To train a well-performed predictive model by taking an object function:

$$\mathcal{O}(y) = \arg\min_{y} \{ L_0 + L_{reg} \} . \tag{1}$$

Here, $L_0$ is the supervised loss from a data-driven perspective, often penalizing the distance between the predicted value $\hat{y}$ and the ground true value $y$ from the machine learning model, and $L_{reg}$ is a differentiable function on model parameters that need additional regularization. Differentiation through argmin using reversed structured, in terms of layer-by-layer, logical inferences to achieve the end-to-end optimization learning. The nested problem shown in Eq.(1) is also known as bilevel optimization problem [39]. In this case, $L_0 + L_{reg}$ is the outer-level optimization problem, and $L_{reg}$ is the inner-level optimization problem. Eq.(1) can be solved using Karush-Kuhn-Tucker (KKT) [40] conditions to constrain the inner-level optimization problem since the $L_0$ is convex with the commonly used cross- entropy loss function.

## 3. OUR METHOD

In this section, we first outline the network construction, along with the graph notations employed. Then, we introduce the objective function, followed by a specific computation detail and a comprehensive explanation of the optimal transport approach. Lastly, we elaborate on the optimization solution that enables the end-to-end training manner of the proposed method.

## 3.1. Network Construction

As depicted in Figure 1, our method consists of two-view GNNs for acquiring modality-specific representations. A modality-wise structure learning (MoSL) layer is used to calculate optimal transport between the two modalities, enabling the learning of node correspondences. A modality fusion layer dedicated to fusing modalities into a unified representation. Finally, the two fully connected layers are designed for denoising the unified representation and executing graph classification.

Specifically, ***Method Input:*** We focus on learning from a single object that possesses two modalities. In each modality, a graph is represented as $\mathcal{G} = (X, \mathcal{L})$, where $X \in \mathbb{R}^{n \times d}$ represents the node feature matrix encoding n nodes with $d$ feature dimensions, and $\mathcal{L} \in \mathbb{R}^{n \times n}$ denotes the corresponding Laplacian matrix. We use the subscripted number to denote the modality index throughout the paper. ***GNNs:*** While each GNN tasks a modality-specific input, it extracts the modality-specific representation through three graph convolutional layers. The model parameters denoted as $\Theta = \{\theta_1, \theta_2, \theta_3\}$, each having an associated hidden dimension size of $\{d_{\theta_1}, d_{\theta_2}, d_{\theta_3}\}$. The convolution operation in GNNs can use the method described in ChebNet [6], or GCN [9]. ***MoSL and Fusion:*** Following the generation of modality-specific representations by GNNs, we employ the learned node feature matrices $Z$ for both MoSL and fusion. In the MoSL layer, we establish node correspondences through the use of an optimal transport metric. This not only determines the most efficient flow of structural learning between corresponding nodes but also quantifies the structural distance between modalities. We delve deeper into this concept by introducing it in our objective function in Section 3.2 and elaborating on its formulation and rationale in Section 3.3. In the modality fusion layer, we apply a fusion technique to combine $Z$ into a unified representation $H$. The fusion technique can be modality-wise max pooling, concatenation, or the Hadamard product, depending on the chosen approach. ***Compatible Learning and Read-out Layer:*** In the final stage, the two consecutives fully connected (FC) layers serve the dual purpose of denoising $H$ and generating a graph-level representation $H'$ for executing the graph classification prediction $\hat{P}$.

## 3.2. Objective of the Method

Parallel engineering and unshared weights of GNNs for modality training pose challenges in modality fusion. These challenges primarily arise from issues such as labeling variability and node correspondence ambiguity. To tackle these issues, we introduce an optimal transport-based MoSL in the network construction during the forward pass. However, how to leverage MoSL to enhance the fusion process becomes a problem. This motivates us to jointly learn graph embedding and MoSL. To achieve this, our approach builds upon the foundation of graph embedding learning, which is based on classifier performance within the MVGNN framework. We expand upon this by involving unsupervised learning on the optimal transport metric between $Z_1$ and $Z_2$ for MoSL. Furthermore, we introduce an adaptive element to improve MoSL, thus AMoSL. This adaptive component refines the structural learning flow based on classifier performance, employing a momentum algorithm.

The performance of the graph classifier is evaluated using the cross-entropy loss, which measures the discrepancy between the ground truth distribution $Y$ of class labels and the predicted probability distribution $\hat{P}$. This loss is computed across $C$ classes as follows:

$$L_0 = -\sum_{c=1}^{C} Y log(\hat{P}) \cdot \tag{2}$$

Throughout the paper, we denote $Y_T$ as the true class label and $\hat{P}_T$ the predicted probability of the true class label for a sample.

On the other hand, the unsupervised learning of AMoSL is defined as:

$$L_{AMoSL}(Z_1, Z_2) = reg \cdot OT(Z_1, Z_2) \ . \tag{3}$$

In Eq.(3), the term reg is adaptive and relies on classifier performance through a momentum algorithm. While we minimize the structural distance between modalities in an unsupervised manner, reg strikes a balance of this distance, ensuring it is neither too small to cause overlapped modalities and miss out on complementary knowledge nor too large to raise labeling variability and disregard shared knowledge. To achieve this balance, we link reg to the classifier's performance, quantified by the probability $\hat{P}_T$ of being the true class. Since $\hat{P}_T$ is an indicator of the accuracy of the unified representation in a classifier. If $\hat{P}_T$ suggests that the current unified representation is accurate (signified by a high value of $\hat{P}_T$), we proceed with graph embedding without imposing strict MoSL constraints. On the contrary, if $\hat{P}_T$ indicates inaccuracies (a high value of $(1 - \hat{P}_T)$), we leverage MoSL to enhance and refine the unified representation for improved performance. Algorithm 1 provides a pseudo-code for computing reg during training epochs $t \in [0, T]$.

As outlined, the fundamental formula of MoSL is the optimal transport metric based on $Z_1$ and $Z_2$,

$$OT(Z_1, Z_2) = \sum_{i \in Z_1} \sum_{j \in Z_2} c_{ij} \tilde{f}_{ij} \ , \tag{4}$$

where $c_{ij}$ is the pairwise distance and $\tilde{f}_{ij}$ is the optimal transport (structural learning) flow between corresponding node $i$ and $j$. We provide a detailed computation of $OT(Z_1, Z_2)$ and its insights regarding MoSL in Section 3.3.

As a result, the objective function of the proposed method is:

$$min \ L \triangleq L_0 + \lambda \cdot L_{AMoSL}(Z_1, Z_2) \ , \tag{5}$$

with $\lambda \in (0, 1)$ being a hyper-parameter.

## 3.3. Optimal Transport-based MoSL

Here, we introduce the computational specifics of $OT(Z_1, Z_2)$ and provide a detailed discussion of its function in the proposed method.

**Computation of $OT(Z_1, Z_2)$.** To calculate the optimal transport-based MoSL between $Z_1$ and $Z_2$, we must consider two key components specific to the problem: the distances between nodes across modalities and the weight assigned to each node. The first step involves establishing node correspondences between modalities.

This process hinges on a distance matrix that represents the dissimilarities between nodes. The dimensions of this matrix depend on the number of nodes present in both modalities. We employ the cosine distance metric to compute $c_{ij}$ in Eq.(4),

$$c_{ij} = 1 - cos(z_{1i}, z_{2j}) \ ,$$
$$= 1 - \frac{z_{1i}^T z_{2j}}{\|z_{1i}\| \|z_{2j}\|} \quad \forall i, j \in \mathbb{R}^n \tag{6}$$

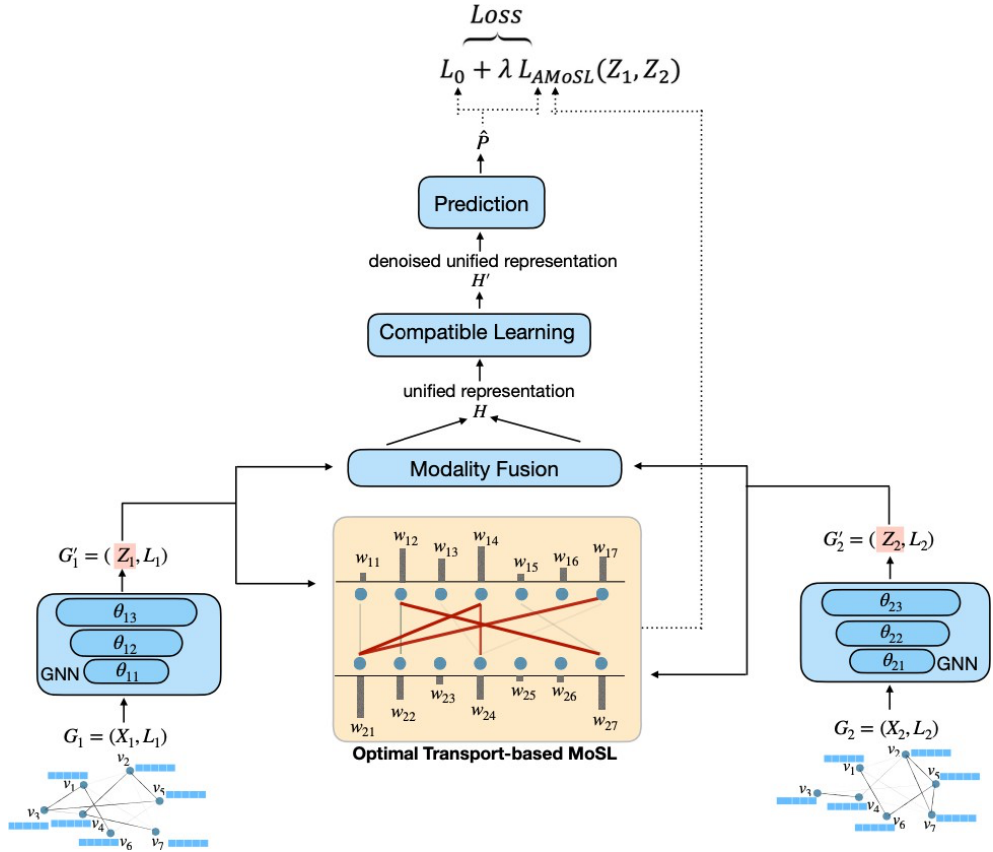

Fig. 1: This figure shows the network construction of AMoSL in MVGNNs with a view number specified to two.

---

**Algorithm 1**: An algorithm to compute *reg*

---

**Input:** Exponential decay rate for the moment estimate, $\gamma \in [0, 1)$

**Output:** reg value at training epoch $t$

1  **for** $t \in [0, T)$ **do**
2      Get the predicted probability of being the true class $\hat{P}_T$ from the $\hat{P}$
3      **if** *time step t = 0* **then**
4          | Compute: $reg \leftarrow (1 - \hat{P}_T)$ ;
5      **else**
6          | Compute: $reg \leftarrow \gamma \cdot reg + (1 - \gamma) \cdot (1 - \hat{P}_T)$ ;
7      **end**
8      t $\leftarrow$ t + 1 ;
9  **end**

---

where $z_{1i} \in \mathbb{R}^{d_{\theta_3}}$ is the $i$-th node feature vector in $Z_1$ and $z_{2j} \in \mathbb{R}^{d_{\theta_3}}$ is the $j$-th node feature vector in $Z_2$. Smaller values of $c_{ij}$ signify the correspondence between nodes $(i, j)$ and the intention of a structural learning flow between them for shared knowledge preservation.

After successfully identifying corresponding nodes using the distance matrix, the next factor to consider is the node weights, which play a crucial role in determining the structural learning flow between these corresponding nodes. To illustrate this, let's consider a practical example: imagine we have movies A, B, and C, with pairs (A, B) and (A, C) representing corresponding movies. Now, when recommending a movie to a user based on their preference for movie A, it makes more sense to prioritize the more popular (or possibly other specific merits) movie if both candidate movies have the same similarity to movie A. Therefore, in cases where distances are equal within a set of corresponding nodes, nodes with higher weights should receive a greater allocation of structural learning flow to their corresponding nodes, particularly if those corresponding nodes also possess higher weights.

Unlike the approach used in [23][31], which assigns a uniform weight to all nodes in node classification tasks, we incorporate two essential factors when determining node weights in graph classification tasks. First, we take into account each node's feature vector. Each node is assigned a feature score (FS) that summarizes its feature vector, defined as $FS = \sum_{d_{\theta_3}} (Z.)$. Second, to facilitate fusion for a more grounded unified representation, we assess each node's contribution score (CS) to the unified representation based on the chosen fusion technique. In the case of,

- modality-wise maximum pooling operation [28], which selects the maximum values element-wise from $Z_1$ and $Z_2$, i.e., $(H)_{r,o} = max\left((Z_1)_{r,o}, (Z_2)_{r,o}\right)$. In this operation, we assign a contribution score of one to the selected element in the unified representation. Consequently, the total contribution score of a node in $Z_1$ and $Z_2$ is computed as follows:

$$\left(CS_{Z_1}\right)_r = \sum_{o=1}^{d_{\theta_3}} 1\left[(Z_1)_{r,o} \geq (Z_2)_{r,o}\right], \quad \forall r \in Z_1,$$

$$\left(CS_{Z_2}\right)_r = \sum_{o=1}^{d_{\theta_3}} 1\left[(Z_1)_{r,o} < (Z_2)_{r,o}\right], \quad \forall r \in Z_2, \tag{7}$$

- concatenation $(H = Z_1 \| Z_2)$ or Hadamard product $(H = Z_1 \odot Z_2)$ operation, where all features are fused into the unified representation. Hence, we assign a contribution score of 0.5 to every element. Consequently, the total contribution score of a node in $Z_1$ and $Z_2$ is equal to $(CS)_r = \sum_{o=1}^{d_{\theta_3}} \frac{1}{2}$. Still, on every feature dimension, $\left(CS_{Z_1}\right)_{ro} + \left(CS_{Z_1}\right)_{ro} = 1$.

Next, we proceed to normalize the score $(CS)_r$ of each node to the value of the feature dimension $d_{\theta_3}$ in the final layer of the GNN:

$$\left(\widehat{CS}\right)_r = \frac{(CS)_r}{d_{\theta_3}} \tag{8}$$

As a result, a node weight is calculated as follows:

$$w = max\{0, FS \times \widehat{CS}\}. \tag{9}$$

Here, we employ the max($\cdot$) function to ensure non-negative weights.

Given the two crucial components, distances c and weights $w$, the optimal transport can determine which nodes should engage in structural learning with one another and establish the magnitude of the structural learning flow between them. This can be accomplished by solving a linear programming problem to obtain the optimal transport (structural learning) flow $\tilde{f}$ that is disclosed in Eq.(4). Consider nodes $i$ and $j$ in $Z_1$ and $Z_2$ with weights $w_i$ and $w_i$ respectively, $\tilde{f}_{ij}$ can be solved as follows:

$$\min_{f_{ij}} \sum_{i \in Z_1} \sum_{j \in Z_2} c_{ij} f_{ij} \tag{10}$$

$$s.t \quad f_{ij} \geq 0, \quad \sum_{i \in Z_1} f_{ij} \leq w_j, \quad \sum_{j \in Z_2} f_{ij} \leq w_i,$$

$$\sum_{i \in Z_1} \sum_{j \in Z_2} c_{ij} f_{ij} = \min\{\sum_{i \in Z_1} w_j, \sum_{i \in Z_1} w_j\} .$$

This gives us, $\tilde{f}_{ij} > 0$ if nodes should engage in structural learning, otherwise $\tilde{f}_{ij} = 0$. After obtaining the optimal structural learning flow $\tilde{f}_{ij}$, the optimal transport between $Z_1$ and $Z_2$ is calculated $OT\ (Z_1, Z_2)$ $= \sum_{i \in Z_1} \sum_{j \in Z_2} c_{ij} \tilde{f}_{ij}$.

**Interpretation of $\boldsymbol{OT\ (Z_1, Z_2)}$.** Let $\Pi = \{(i,j) \mid \tilde{f}_{ij} > 0\}$ be the set of corresponding nodes that should engage in structural learning through the solution of $OT\ (Z_1, Z_2)$. The unsupervised learning objective of $OT\ (Z_1, Z_2)$ in Eq. (5) is aiming at minimizing the distances among the corresponding nodes in $\Pi$. This is because only $(i,j) \in \Pi$ exhibit meaningful structural learning flows, as $\tilde{f}_{ij} > 0$.

Moreover, minimizing $OT\ (Z_1, Z_2)$. can resolve the issue of two concerns in existing MVGNNs. First, as $Z_1$ and $Z_2$ are the inputs of OT, minimizing $OT\ (Z_1, Z_2)$ is equivalent to the minimization of distance between $Z_1$ and $Z_2$. This offers a resolution to the issue of the two modalities falling apart within a given feature space. Second, minimizing the distances among the nodes in $\Pi$ implicitly mitigates the challenge of potential loss of shared knowledge between modalities.

## 3.4. End-to-end Training

The objective function in Eq. (5) is a bilevel optimization problem, where the outer-level optimization problem is to optimize model parameters of the graph neural network, and the inner-level one is to optimize the structural learning flow between corresponding nodes for computing the optimal transport. To optimize this challenging problem, we design an end-to-end training strategy that efficiently solves it using the stochastic gradient descent algorithm. Since it is a convex optimization problem, we can solve it with an existing convex solver efficiently in the forward pass. However, it is challenging to compute the gradient in the backward pass. In particular, when computing the gradient of the loss function $L$ with respect to the model parameter $\theta$, we need to compute $\frac{\partial \tilde{f}(\theta)}{\partial \theta}$. Therefore, to enable the end-to-end training, we need compute $\frac{\partial \tilde{f}}{\partial \theta}$ efficiently.

To address this challenge, we construct the Lagrangian function for Eq. (10) as follows:

$$\mathcal{T}(f(\theta), \eta, \nu) = c^T f(\theta) + \eta^T P(f(\theta)) + \nu^T Q(f(\theta)), \tag{11}$$

where $\eta > 0$ and $\nu$ are dual variables, $P(f(\theta))$ represents the inequality constraints, and $Q(f(\theta))$ denotes the equality constraints. Then, according to the KKT condition, i.e., $G \triangleq \mathcal{T}(\tilde{f}(\theta), \tilde{\eta}, \tilde{\nu})$ where $\tilde{f}(\theta), \tilde{\eta},$ and $\tilde{\nu}$ denotes the optimal solution of Eq.(11), it is easy to obtain $\frac{\partial \tilde{f}}{\partial \theta}$ by taking the gradient of $G$ with respect to $\theta$, which is shown as follows:

$$\frac{\partial \tilde{f}(\theta)}{\partial \theta} = -\left(\frac{\partial G}{\partial \tilde{f}(\theta)}\right)^{-1} \frac{\partial G}{\partial \theta} \tag{12}$$

where the right-hand side is easy to compute based on Eq. (11) (See Eqs.(9-10) in [41]). As a result, by plugging this step into the backpropagation procedure, we train our graph neural network in an end-to- end manner. In essence, we enable AMoSL awareness before the fusion layer, tackle the fusing challenges, and implicitly enhance the unified representation for better performance in the downstream task.

## 4. EXPERIMENTS

To evaluate the effectiveness of our proposed optimal transport-based AMoSL approach in MVGNNs, we conduct comprehensive experiments using six benchmark datasets for graph classification. This section begins by providing a summary of dataset information and implementation details for our experiments. Subsequently, we perform ablative studies to validate each component within our method design. Finally, we present a comparative analysis of our proposed method against state-of-the-art approaches across the six benchmark datasets.

### 4.1. Experimental Setup

**Datasets.** We use six benchmark datasets for graph classification from TUDatasets [42]. In our dataset selection, we span diverse domains, including the recognition of small molecular networks such as MUTAG, BZR_MD, PTC_MR, and ER_MD. We also delve into the realm of computer vision by considering the recognition of Cuneiform signs, and we explore brain disease prediction, as exemplified by KKI. Detailed statistics for these datasets are provided in Table 1.

In dataset preparation, we begin with single-modality datasets $\mathcal{G} = (X, A)$. We use them as the first modality $\mathcal{G}_1 = (X, \mathcal{L}_1(A))$. To create the second modality, we take a different approach compared to techniques like node permutation [23] or the addition of noise [26], which are commonly used in MVGNNs for node classification tasks. Instead, we generate a diverse graph topology $\mathcal{S}$ based on the Mahalanobis distances matrix ($\mathcal{D} \in \mathbb{R}^{n \times n}$) between node features with a randomly generated transformation matrix ($M \in \mathbb{R}^{d \times d}$), and normalize it by a standard Gaussian distribution [16]. Such that, $\mathcal{S} = exp(-\mathcal{D}/2)$, with $d(x_i, x_j) = \sqrt{(x_i - x_j)^T M(x_i - x_j)}$, for all $d(x_i, x_j) \in \mathcal{D}$. As a result, the second input graph $\mathcal{G}_2 = (X, \mathcal{L}_2(A, \mathcal{S}))$ is constructed, featuring a semi-synthetic graph topology.

Table 1: Statistics of the Datasets

| Datasets | Graphs | Class | Avg. Vertices | Features |
|----------|--------|-------|---------------|----------|
| MUTAG | 188 | 2 | 17.93 | 7 |
| BZR_MD | 306 | 2 | 21.30 | 8 |
| PTC_MR | 344 | 2 | 14.29 | 18 |
| ER_MD | 446 | 2 | 21.33 | 10 |
| Cuneiform | 267 | 30 | 21.27 | 3 |
| KKI | 83 | 2 | 26.96 | 190 |

**Implementation Details and Configurations Input data.** We adopt a data partitioning approach following prior studies [16][43][44]. Our methodology incorporates a 10-fold cross-validation technique. During training and evaluation of our proposed method, we utilize a mini-batch size of 32.

The data preprocessing steps are consistent with those outlined in Section 4.1.

**Network Configuration.** Network Configuration: In line with the network architecture discussed in Section 3.1, each Graph Neural Network (GNN) in our model consists of three graph convolutional layers. These layers have embedding dimensions set as $d_{\theta_1} = 16$, $d_{\theta_2} = 64$, $d_{\theta_3} = 128$. ReLU activation functions and dropout layers with a rate of 0.1 follow each convolutional layer. We employ a convolution operation as described in ChebNet [6], with a Chebyshev degree ($K$) of 6, or in GCN [9]. The fusion layer utilizes modality-wise max pooling to generate a unified representation. In the compatibility learning layer, we apply a linear layer with an embedding dimension of 128, followed by a ReLU activation layer and a dropout layer with a rate of 0.1. The read-out layer employs graph-level max pooling among the nodes to create the graph-level representation for class prediction.

**Parameter Tuning.** We tune the moment estimate parameter $\gamma$ in Algorithm 1 over the range {0.01, 0.02, ..., 0.99, 1.}, and the hyperparameter $\lambda$ in Eq.(5) over the values {5e-2, 1e-3, 5e-3 ..., 1e-5, 5e-5}.

**Training Configuration.** We evaluate the method's performance using classification accuracy as the primary metric. We employ the Adam optimizer with a learning rate of 5e-3 and train for a total of 200 epochs (T = 200). To solve the Linear Programming (LP) problem in the optimal transport metric, we leverage the GPU-accelerated convex optimization solver QPTH [45] and compute gradients during the backward pass.

## 4.2 Method Analysis

In this section, we implement various experiments to evaluate the effectiveness of our method design by exploring multiple design variants and comparing the performances.

**Comparison of alternative distance metrics for AMoSL.** Given that optimal transport indeed serves as a distance metric, we conducted a performance comparison of AMoSL when employing alternative distance metrics, including Manhattan distance, Euclidean distance, and cosine distance, as replacements for the optimal transport metric in Eq.(4). It is important to emphasize that these alternative distance metrics assume a constant graph topology, which leads them to solely measure the distance between two modalities without accounting for the identification of corresponding nodes. Furthermore, these alternatives operate under the assumption of a uniform structural learning flow between nodes, without considering feature importance or the significance of contribution to the unified representation. As depicted in Table 2, none of the alternative distance methods demonstrated superior classification performance compared to optimal transport. This comparison underscores the effectiveness of optimal transport as the metric for AMoSL because of its capacity to identify corresponding nodes and allocate the optimal structural learning flow between them.

Table 2: Comparison of different distance metrics

| Datasets | Distance Metrics | | | |
|---|---|---|---|---|
| | Manhattan | Euclidean | cosine | Optimal Transport |
| MUTAG | 88.9 ± 5.4 | 88.9 ± 4.9 | 88.9 ± 4.9 | **91.0 ± 4.1** |
| BZR_MD | 80.0 ± 5.9 | 81.1 ± 6.8 | 81.1 ± 6.4 | **81.7 ± 5.8** |
| PTC_MR | 67.8 ± 5.0 | 71.7 ± 6.2 | 72.0 ± 5.1 | **75.3 ± 7.9** |
| ER_MD | 78.1 ± 3.1 | 79.2 ± 5.3 | 79.4 ± 3.9 | **80.1 ± 2.9** |
| Cuneiform | 83.5 ± 11.8 | 83.9 ± 9.1 | 85.1 ± 10.3 | **86.5 ± 8.3** |
| KKI | **83.3 ± 14.3** | 82.2 ± 10.2 | 80.0 ± 10.9 | **83.3 ± 10.2** |

**The Influence of the Adaptive Effect designed in the Objective Function.** We also investigate the influence of the adaptive effect designed in the objective function, i.e., the reg term in Eq.(3). As discussed in Section 3.2, the proposed optimal transport-based AMoSL is adapted to the classifier's performance. However, we also explore a radical scenario where the adaptive effect is eliminated by setting the regularization parameter *reg* = 1, allowing unconditional unsupervised learning on structure distance. Table 3 presents the results of structure distance and model performance under this radical scenario. In this experiment, the network is set up to use the ChebNet graph convolution operation with $K = 1$, while other configurations remain consistent with those described in Section 4.1. When structure distance learning is always allowed, we observe significantly small structure distances, which excessively emphasize the importance of minimizing distances between modalities. This converges the modalities too close in the feature space and hinders MVGNNs from utilizing complementary knowledge from modalities. As a result, none of the radical scenarios can train classifiers that outperform the proposed method. This underscores the significance of maintaining a balance between modalities, thus emphasizing the importance of the adaptive design in the proposed method.

Table 3: Evaluation of the adaptive effect. We use ✗ to represent *reg* = 1, for methods without the adaptive effect, and ✓ to represent methods with the adaptive effect.

| Datasets | Fusion Techniques | Distance | | Accuracy | |
|---|---|---|---|---|---|
| | | ✗ | ✓ | ✗ | ✓ |
| MUTAG | max | 3.75 | 14.73 | 89.4 ± 6.2 | **91.0 ± 4.1** |
| | concat | 4.87 | 11.35 | 91.0 ± 5.8 | **91.5 ± 4.8** |
| BZR_MD | max | 4.00 | 14.14 | 81.1 ± 6.4 | **81.7 ± 5.8** |
| | concat | 4.44 | 12.55 | 81.1 ± 6.0 | **82.4 ± 7.0** |
| PTC_MR | max | 24.25 | 29.96 | 72.0 ± 5.5 | **75.3 ± 7.9** |
| | concat | 4.76 | 37.84 | 71.1 ± 4.7 | **71.6 ± 6.0** |
| ER_MD | max | 3.19 | 13.39 | 78.3 ± 4.4 | **80.1 ± 2.9** |
| | concat | 3.75 | 11.32 | 78.7 ± 4.2 | **79.6 ± 5.0** |
| Cuneiform | max | 24.92 | 30.83 | 84.6 ± 11.3 | **86.5 ± 8.3** |
| | concat | 20.11 | 14.70 | 86.5 ± 6.5 | **88.0 ± 8.3** |
| KKI | max | 13.96 | 35.86 | 82.2 ± 12.4 | **83.3 ± 10.2** |
| | concat | 11.26 | 37.35 | 84.4 ± 11.3 | **85.6 ± 13.2** |

**AMoSL incorporates different fusion techniques.** The fusion layer in MVGNNs plays a pivotal role in fusing representations from different modalities, enabling the method to exploit complementary and shared knowledge from different modalities. The resulting unified representation can be more informative than the individual modality-specific representations for the downstream tasks. In the context of our proposed method, we use a modality-wise max pooling technique, which selects the element-wise max present among modalities to the unified representation. We find other fusion techniques, such as concatenation, which effectively combines all representations along the feature dimension, and Hadamard product, which multiplies all representations together, are widely used in the domain. In this study, we do not seek to establish the correctness of using the max pooling technique but to understand how our approach encompasses alternative fusion techniques. As evident from the results presented in Table 4, our approach outperforms MVGNNs without AMoSL, suggesting AMoSL alongside different fusion techniques in MVGNNs can enhance the unified representation to a more accurate classifier.

Table 4: Comparison of different fusion techniques. We compare methods with/without the proposed optimal transport-based AMoSL under different fusion techniques. The performance differences that exceed 1% are denoted with a single asterisk (*), while those surpassing the 2% threshold are marked with a double asterisk (**).

| Datasets | With AMoSL? | Fusion Techniques | | |
|---|---|---|---|---|
| | | max-pooling | concatenation | Hadamard Product |
| MUTAG | ✗ | 93.62 ± 4.58 | 91.52 ± 5.33 | 94.15 ± 4.96 |
| | ✓ | 94.74 ± 4.70* | 93.62 ± 5.15** | 94.15 ± 4.37 |
| BZR_MD | ✗ | 72.16 ± 4.67 | 72.85 ± 7.35 | 71.87 ± 7.88 |
| | ✓ | 74.79 ± 4.50** | 75.19 ± 6.60** | 74.91 ± 6.44** |
| PTC_MR | ✗ | 80.56 ± 4.01 | 79.85 ± 2.96 | 78.03 ± 4.15 |
| | ✓ | 81.60 ± 3.43* | 81.65 ± 4.18* | 80.56 ± 3.76** |
| ER_MD | ✗ | 81.38 ± 7.32 | 80.73 ± 5.78 | 79.71 ± 5.20 |
| | ✓ | 82.62 ± 7.50* | 83.31 ± 7.01** | 82.02 ± 5.61** |
| Cuneiform | ✗ | 88.33 ± 6.20 | 88.75 ± 5.25 | 87.22 ± 6.75 |
| | ✓ | 89.49 ± 5.96* | 89.86 ± 5.30* | 88.75 ± 5.99* |
| KKI | ✗ | 81.11 ± 12.20 | 81.11 ± 14.95 | 82.22 ± 11.33 |
| | ✓ | 86.67 ± 12.96** | 83.33 ± 12.42** | 87.78± 10.48** |

## 4.3. Comparison with State-of-the-art Methods

The experimental results for graph classification are presented in Table 5. We observe that our method consistently outperforms all baseline methods across all datasets. In the following, we analyze the results in two scenarios, providing further evidence for the effectiveness of the proposed method.

**Comparison with single-view methods.** First, we evaluate the performance of ChebNet and GCN, both renowned as foundational methods in the GNN domain. Their popularity arises from the simplicity of construction and efficient learning, especially in the case of GCN. Our findings demonstrate that the classification accuracy achieved by ChebNet and GCN in learning either the 1st or 2nd modality comparable with that of other methods, namely GIN and InfoGraph when handling the 1st modality. This highlights our rationale for selecting ChebNet and GCN as the base methods in the GNN component of our proposed method.

Second, we compare our method with single-view methods. Our observations indicate that the classifiers trained by our methods, ChebNet-MVGNN+AMoSL and GCN-MVGNN+AMoSL, consistently outperform those of the single-view methods. This evidence strongly suggests that a unified representation, as trained by our approach, is better for meeting downstream tasks compared to modality-specific single-view representations.

**Comparison with multi-view methods.** To demonstrate the effectiveness of the proposed Adaptive Modality-wise Structural Learning (AMoSL), we conduct a comparative analysis against other Multi-View Graph Neural Networks (MVGNNs), namely Multigraph and MVAGC. We categorize the compared methods into two groups: ChebNet-like and GCN-like, aligning with the respective graph convolution operation employed. Our results consistently indicate that AMoSL outperforms both Multigraph and MVAGC. This performance improvement ranges from 0.4% in the Cuneiform dataset using GCN-like methods to a substantial 5% increase in classification accuracy for the KKI dataset using ChebNet-like methods. These findings provide strong evidence that AMoSL significantly enhances the unified representation compared to methods lacking this adaptive modality-wise structural learning component.

It's noteworthy that most existing studies in this domain predominantly focus on brain networks, and unfortunately, we lacked access to domain-specific data and code resources to perform a comprehensive comparison in that specific context.

Table 5: Graph Classification Results in terms of 10-fold Average Accuracy

| Method | Datasets | | | | | |
|---|---|---|---|---|---|---|
| | MUTAG | BZR_MD | PTC_MR | ER_MD | Cuneiform | KKI |
| GIN (1st modality)[12] | $90.5 \pm 7.4$ | $66.3 \pm 11.3$ | $71.5 \pm 8.2$ | $68.1 \pm 6.1$ | - | $80.6 \pm 17.6$ |
| InfoGraph (1st modality)[46] | $91.4 \pm 0.1$ | $79.4 \pm 0.1$ | $72.1 \pm 0.1$ | $78.2 \pm 0.1$ | $88.4 \pm 0.1$ | $65.6 \pm 0.2$ |
| ChebNet (1st modality) [6] | $91.5 \pm 5.8$ | $78.7 \pm 6.5$ | $70.4 \pm 5.0$ | $78.0 \pm 4.4$ | $87.0 \pm 5.1$ | $78.9 \pm 12.6$ |
| ChebNet (2nd modality) [6] | $92.6 \pm 4.8$ | $79.0 \pm 8.7$ | $69.0 \pm 4.9$ | $76.9 \pm 4.6$ | $87.9 \pm 5.1$ | $81.1 \pm 13.2$ |
| ChebNet-Multigraph [43] | $93.1 \pm 4.7$ | $81.6 \pm 7.2$ | $73.6 \pm 2.7$ | $79.6 \pm 4.2$ | $88.7 \pm 6.0$ | $83.3 \pm 11.4$ |
| ChebNet-MVAGC [16] | $93.6 \pm 4.6$ | $81.4 \pm 7.3$ | $72.2 \pm 4.7$ | $80.6 \pm 4.0$ | $88.3 \pm 6.2$ | $81.1 \pm 12.2$ |
| ChebNet-MVGNN+AMoSL | $\mathbf{94.7 \pm 4.7}$ | $\mathbf{82.6 \pm 7.5}$ | $\mathbf{74.8 \pm 4.5}$ | $\mathbf{81.2 \pm 3.4}$ | $9.5 \pm 6.0$ | $\mathbf{86.7 \pm 12.9}$ |
| GCN (1st modality) [9] | $89.9 \pm 4.9$ | $79.3 \pm 6.5$ | $73.4 \pm 6.8$ | $79.8 \pm 4.2$ | $88.7 \pm 6.5$ | $82.2 \pm 11.3$ |
| GCN (2nd modality) [9] | $89.9 \pm 4.9$ | $78.7 \pm 6.8$ | $72.6 \pm 7.8$ | $79.8 \pm 3.1$ | $88.7 \pm 5.6$ | $78.9 \pm 16.1$ |
| GCN-Multigraph [43] | $91.5 \pm 3.4$ | $79.4 \pm 7.3$ | $74.3 \pm 4.7$ | $79.0 \pm 4.0$ | $88.4 \pm 5.4$ | $81.1 \pm 12.2$ |
| GCN-MVAGC [16] | $91.0 \pm 4.1$ | $79.7 \pm 5.4$ | $72.2 \pm 5.6$ | $81.0 \pm 4.1$ | $89.1 \pm 5.4$ | $84.4 \pm 10.2$ |
| GCN-MVGNN+AMoSL | $\mathbf{91.5 \pm 4.2}$ | $\mathbf{82.7 \pm 6.5}$ | $\mathbf{74.6 \pm 7.3}$ | $\mathbf{82.5 \pm 3.3}$ | $\mathbf{89.5 \pm 4.9}$ | $\mathbf{85.6 \pm 10.0}$ |

## 5. CONCLUSION

This paper introduces a novel approach for Multi-view Graph Neural Networks (MVGNNs), centered around optimal transport-based AMoSL. It encompasses both network construction and objective function design, aiming to tackle the complex challenges posed by fusing, particularly in scenarios with unclear inter-modality correlations, in terms of labeling variability and node correspondence neglect. Our approach jointly trains graph embedding and MoSL, leveraging an efficient optimization technique that facilitates end-to-end training. We conduct a comprehensive evaluation of the design within our proposed method, and the results demonstrate its superiority in training improved graph classifiers compared to existing approaches.

## ACKNOWLEDGEMENTS

The authors would like to thank the anonymous reviewers for their insightful comments. This work is partially supported by the National Science Foundation (NSF) Award Nos. 2311758 and 2134203. Some of the experiments are also conducted on the NSF Chameleon cloud. Any opinions, findings and conclusions or recommendations expressed in this material are those of the author(s) and do not necessarily reflect the views of the National Science Foundation.